\documentclass[letterpaper]{article} 
\usepackage{aaai25}
\usepackage{times}  
\usepackage{helvet}  
\usepackage{courier}  
\usepackage[hyphens]{url}  
\usepackage{graphicx} 
\usepackage{natbib}  
\usepackage{caption} 
\usepackage{algorithm}
\usepackage{algorithmic}

\usepackage{amsfonts}
\usepackage{amsmath}
\usepackage{multirow}
\usepackage{subfigure}

\usepackage{newfloat}
\usepackage{listings}

\DeclareCaptionStyle{ruled}{labelfont=normalfont,labelsep=colon,strut=off} 
\floatstyle{ruled}
\newfloat{listing}{tb}{lst}{}
\floatname{listing}{Listing}

\title{TB-HSU: Hierarchical 3D Scene Understanding with Contextual Affordances}
\author{
    Wenting Xu\textsuperscript{\rm 1},
    Viorela Ila\textsuperscript{\rm 2},
    Luping Zhou\textsuperscript{\rm 1},
    Craig T. Jin\textsuperscript{\rm 1}\\
}
\affiliations{
    \textsuperscript{\rm 1}School of Electrical and Computer Engineering, The University of Sydney\\

     \textsuperscript{\rm 2}School of Aerospace, Mechanical and Mechatronic Engineering, The University of Sydney\\
     \{wenting.xu, viorela.ila, luping.zhou, craig.jin\}@sydney.edu.au
}

\usepackage{bibentry}

\begin{document}
\maketitle



\begin{abstract}
The concept of function and affordance is a critical aspect of 3D scene understanding and supports task-oriented objectives. In this work, we develop a model that learns to structure and vary functional affordance across a 3D hierarchical scene graph representing the spatial organization of a scene. The varying functional affordance is designed to integrate with the varying spatial context of the graph. More specifically, we develop an algorithm that learns to construct a 3D hierarchical scene graph (3DHSG) that captures the spatial organization of the scene. Starting from segmented object point clouds and object semantic labels, we develop a 3DHSG with a top node that identifies the room label, child nodes that define local spatial regions inside the room with region-specific affordances, and grand-child nodes indicating object locations and object-specific affordances. To support this work, we create a custom 3DHSG dataset that provides ground truth data for local spatial regions with region-specific affordances and also object-specific affordances for each object. We employ a Transformer Based Hierarchical Scene Understanding model (TB-HSU) to learn the 3DHSG. We use a multi-task learning framework that learns both room classification and learns to define spatial regions within the room with region-specific affordances. Our work improves on the performance of state-of-the-art baseline models and shows one approach for applying transformer models to 3D scene understanding and the generation of 3DHSGs that capture the spatial organization of a room. The code and dataset are publicly available.
\end{abstract}

%
\begin{links}
    \link{Code, Dataset}{Github.com/WentingXu3o3/TB-HSU}
\end{links}


\begin{figure}[t]
    \centering
    \includegraphics[width=\columnwidth]{./figures/3Layers.pdf}
    \caption{\textbf{Overview:}
    The affordances of objects vary depending on the granularity of the context. We propose a multi-task method, Transformer Based Hierarchical Scene Understanding (TB-HSU), employing a transformer-based model for 3D scene understanding to classify the room and identify the region, forming a 3D Hierarchical Scene Graph (3DHSG) with three layers: Objects, Regions, and Rooms.}
    \label{fig:3_mesh_layers}
\end{figure}

\section{Introduction}
\label{sec:intro}
Understanding visual scenes is fundamental to machine perception, requiring not only accurate object representation, but also consideration of how these objects might be arranged in their functional zones. In this work, we take a navigational and task-oriented perspective to scene understanding to assist robots in deciding which areas to explore to complete a task. In this light, functional considerations require appropriate representation and consideration of the potential usage of objects within the scene, often referred to as object affordances.  The annotation of affordances can be labor-intensive and thus many 3D scene datasets~\cite{chang_matterport3d_2017,dai_scannet_2017,straub_replica_2019} lack affordance annotations for objects, with only a few, such as AI2-THOR~\cite{kolve_ai2-thor_2022}, 3DSSG~\cite{wald_learning_2020} and 3D AffordanceNet~\cite{deng_3d_2021} providing affordance annotations. In AI2-THOR~\cite{kolve_ai2-thor_2022}, a fixed affordance list with actionable attributes is assigned to each object class, such as ``ArmChair - Receptacle, Moveable" and ``Book - Openable, Pickupable". The 3DSSG~\cite{wald_learning_2020} dataset similarly defines fixed affordance annotations with actionable attributes for each object class, with examples such as ``Armchair - Lying on, Sleeping, Moving" and ``Book - Reading, Storing in, Storing on." Additionally, 3D AffordanceNet~\cite{deng_3d_2021} takes a different approach and annotates specific parts of the object with affordance labels. For example, for a mug, different vertices are associated with affordances such as ``pour, wrap, contain''. However, all current datasets with affordance annotations provide either a fixed affordance list or vary the affordance based on the particular physical contact points without considering how affordances change depending on context.

In this work, we assume that the affordances of objects should vary depending on the spatial context, i.e., the precision and extent of the surroundings and location. For example, chairs placed beside a table and the table situated next to a kitchen stove primarily serve the purpose of dining. ``Narrowing down'' or ``zooming in'' on the region around the chairs and table would indicate the chair serves as a place for sitting, and the table serves as a surface for placing items. Similarly, chairs around a sofa facing a TV are primarily used for resting, but when we ``zoom in'' to the region around the chairs and sofa, the chair and the sofa both serve the purpose of sitting. Continuing in this fashion, a bed with a pillow within a room primarily serves the purpose of sleeping, while narrowing down on the bed and pillow indicates that the pillow serves the purpose of propping the head or body, while the bed serves the purpose of lying on. A pillow on an armchair serves the purpose of resting; ``zooming in'' on the pillow indicates the purpose of propping, while the armchair indicates the purpose of sitting. While the changes in affordances are transparent and obvious to human observers, these transitions in the affordance of objects have not really been addressed in existing 3D scene datasets.

\subsection{Contributions}
\hspace{8pt}\textbf{Firstly}, we introduce the \textbf{3D} \textbf{H}ierarchical \textbf{S}cene \textbf{G}raph (3DHSG) dataset that extends the 3DSSG dataset~\cite{wald_learning_2020}, which itself extends the 3RScan dataset~\cite{wald_rio_2019}. 3DHSG captures the spatial organization for a 3D scene in a three-layered graph, where nodes represent objects, regions within rooms, and rooms. Object nodes include context-specific affordances, while region nodes cluster objects with the same region-specific affordances, and room nodes contain the type of room.

\textbf{Secondly}, we develop a \textbf{T}ransformer \textbf{B}ased \textbf{H}ierarchical \textbf{S}cene \textbf{U}nderstanding (TB-HSU) model, which automatically constructs the 3DHSG for a room using instance-segmented point cloud data and object semantic labels within a multi-task learning framework. The TB-HSU model generates a hierarchical 3D scene graph that captures the spatial organization of a room. 

\textbf{Thirdly}, we validate our proposed TB-HSU method on both our custom 3DHSG dataset and two public benchmarks and demonstrate our promising performance over multiple baseline models consistently. We also showcase how 3DHSG contributes to improving GPT-4o's capacity in question answering, \textit{e.g.} finding an object not in the scene.

\section{Related Work}
\subsection{3D Scene Graphs} Current work often uses hierarchical 3D Scene Graphs~\cite{hughes_hydra_2022,hughes_foundations_2023,rosinol_kimera_2021,rosinol_3d_2020,kim_3-d_2020,wald_learning_2020,scenegraphfusion} to represent information such as object relationships within 3D scenes. A 3D scene graph is a layered graph where nodes represent spatial concepts at multiple levels of abstraction (from objects to regions and rooms, etc.), and edges represent relationships between these concepts (\textit{e.g.}, Figure~\ref{fig:3_mesh_layers}). Armeni et al.~\cite{armeni_3d_2019} pioneered the use of 3D scene graphs in comprehensive semantic scene understanding and proposed the first algorithms to construct a graph that includes the semantics of objects, rooms, and cameras for each layer.  The Hydra model~\cite{hughes_hydra_2022,hughes_foundations_2023}  and SceneGraphFusion model~\cite{scenegraphfusion} introduced a 3D hierarchical scene graph for indoor environments from sensor data. The SceneGraphFusion model incrementally builds a semantic scene graph simultaneously while performing 3D mapping. The Hydra model incrementally builds a 3D hierarchical scene graph in real-time that captures the spatial organization of a scene, e.g., it creates a graph with layers such as buildings, rooms, places, objects and agents, and a semantic 3D mesh. Object segmentation, identification, and relationships are a significant focus for many 3D scene graph models. The affordance of objects is often considered a node attribute in these 3D scene graphs. For example, in~\cite{wald_learning_2020}, 3D semantic scene graphs are created where nodes contain a hierarchy of object classes and also object affordances. In this work, we focus on affordances and consider affordance as a concept that varies with spatial organization and that can assist with solving the 3D scene understanding problem. Our focus stems from a navigational and task-oriented perspective in which varying the affordance with spatial organization can assist with decision-making. 

In this work, we follow the tradition of using hierarchical 3D scene graphs to represent indoor environments with varying levels of spatial organization (e.g. 
objects, regions, and rooms) in a three-layer 3D scene graph (see Figure~\ref{fig:3_mesh_layers}).

\subsection{LLMs and VLMs In 3D Scene Reasoning}
The success of large language models (LLMs) such as GPT~\cite{openai_gpt-4_2024_short} and LLaMa~\cite{touvron_llama_2023} and vision-language models (VLMs) such as CLIP~\cite{radford_learning_2021} has inspired their application to 3D scenes~\cite{hong_3d-llm_2023,gu_conceptgraphs_2023,jatavallabhula_conceptfusion_2023,maggio_clio_2024}. For example, OpenScene~\cite{peng_openscene_2023} embeds dense 3D point features with image pixels and text to answer open-vocabulary queries. An instance of a query can be to find points that are conceptually associated with words such as ``soft'', ``kitchen'', or ``work''. ConceptFusion~\cite{jatavallabhula_conceptfusion_2023} creates a 3D map with ``pixel-aligned'' features that permit multiple kinds of open queries. These multimodal queries can relate to arbitrary concepts, including affordance-related questions, and can return spatial regions consistent with the query. Moreover, ConceptGraph~\cite{gu_conceptgraphs_2023} builds open-vocabulary 3D scene graphs emphasizing spatial relationships between objects through reasoning with LLMs. Their scene graphs can interface with LLMs to give useful facts to robots about surrounding objects's traversability and utility. Consider also the real-time robotics algorithm Clio~\cite{maggio_clio_2024}, designed to build task-driven 3D scene graphs with embedded open-set semantics. Clio forms task-relevant clusters of object primitives based on a set of natural language tasks, clustering the scene into task-relevant semantic regions such as ``Kitchenette", ``Workspace," and ``Conference Room." We find our ideas most similar to Clio, which also considers semantic concepts as varying in ``granularity''. For Clio, the focus is task-driven with the objective of creating 3D scene graphs that retain task-relevant objects and regions only. This approach leaves users or robots with the issue of determining a suitable task list for navigational purposes, which could be particularly difficult when they first encounter a new environment. We certainly agree with the perspective that object affordances should be considered from a task-dependent framework but suggest that object affordances should relate to spatial context as well. We suggest that a 3D hierarchical scene graph representing spatial organization with varying affordances can be used for exploratory as well as task-driven purposes and will likely be useful in queries to LLMs and VLMs.

\section{Methods}
\subsection{The 3DHSG Dataset}
\subsubsection{Overview of the 3DHSG Dataset}
In order to explore the machine analysis of 3D scenes based on the affordances of objects at varying levels of spatial context, we create a custom dataset referred to as the hierarchical scene graph (3DHSG) dataset. To be clear, we take a navigational framework for the classification of the affordance of objects. While the juxtaposition of navigation and object affordance may seem odd, this approach stems from the fact that a robot with a particular task objective in focus will need to consider that within a building, there are different rooms serving differing and multiple purposes and holding objects of varying affordances. In this way, a decision can be made about which rooms to approach based on a room affordance level to evaluate the potential for task completion. Further, upon reaching a particular room, the robot will need to evaluate whether local regions within the room provide the possibility for task completion. This evaluation would be based on more specific aspects of object affordances that match the level of detail associated with the local region of the room. In this way, we integrate a navigational approach into solving task objective problems that systematically refine the spatial and functional information, with the understanding that appropriate object affordance data should be available at each stage to evaluate the likelihood of task completion. Depending upon whether the navigational information is available as stored data or requires direct exploration determines whether the robot must explore the area directly.

To better understand the dataset, it is helpful for us to begin by describing three of its aspects: (i) the data available, (ii) the two machine learning tasks associated with the dataset, and (iii) the development of a hierarchical scene graph (3DHSG) capturing the spatial organization of the scene. The top node of the graph is simply a building that consists of a collection of rooms. A room is then characterized by collections of point clouds that have object labels, and Cartesian coordinate data available. Once a room has been selected, the first machine-learning task is room classification based on the objects within the room. A second machine learning task is to then divide the room into local spatial regions that define region-specific affordances. These local spatial regions form the nodes for the second level of the 3DHSG. The first layer of the 3DHSG consists of nodes representing the individual objects (and their locations) with object-specific affordances. In a sense, we solve the two machine-learning tasks in order to develop the 3DHSG. 

To complete the descriptive overview of the dataset, we describe the data that are stored with the object nodes. There are two levels of affordance data: object-specific and region-specific. These two levels of object affordances are stored in a two-component vector associated with each object. Each object node also contains a list of rooms that are commonly associated with the object.

\subsubsection{Details of the 3DHSG Dataset}
An outline sketch of the custom 3DHSG dataset is provided in Figure~\ref{fig:3_mesh_layers}. To create this dataset, we start with the real-world 3D scene dataset 3RScan~\cite{wald_rio_2019}, which comprises 1482 3D reconstructions of 478 natural indoor environments. Instance-segmented point cloud data and object semantic labels are provided with the 3RScan dataset, while object attributes come from 3DSSG~\cite{wald_learning_2020}. For a given room, we describe the room category and provide each object with a region-specific affordance and object-specific affordance. Objects with the same region-specific affordance form a local region. Further, we also provide a room list for each object indicating the rooms in which it is usually found. The data are then organized into a 3DHSG as described above. 

In order to determine both the region-specific and object-specific affordance data along with the room category, we use the descriptive data provided by GPT~\cite{openai_gpt-4_2024_short}, ConceptNet~\cite{speer_conceptnet_2018} and 3DSSG~\cite{wald_learning_2020}. These descriptive data for the object affordances were then evaluated manually by humans for their suitability for region-specific and object-specific affordances. Specifically, we conducted initial region groups based on the intersections of objects' bounding boxes and then manually refined the groups. The manually selected affordance data for each object was stored as a tuple with two elements: $<$region-specific affordance, object-specific affordance$>$.
For our dataset, we define twelve different room types, 27 different region-specific affordances, and 87 different object-specific affordances. More details are provided in the Appendices.

Similarly to \cite{hughes_foundations_2023},  the graph in our 3DHSG is defined as $\mathcal{G} = (\mathcal{N}, \mathcal{E})$,  where the set of nodes $\mathcal{N}$ can be partitioned into $\ell = 3$ layers. Specifically, $\mathcal{N} = \bigcup_{i=1}^{\ell} \mathcal{N}_i$, where the lowest layer, $\mathcal{N}_1$, describes the objects; the next layer, $\mathcal{N}_2$, describes the regions; and the top layer, $\mathcal{N}_3$, describes the rooms. Each node $n \in \mathcal{N}_i$ in layer $i$ has only a single parent, meaning it shares an edge with at most one node in the layer $\mathcal{N}_{i+1}$ above. This reflects that each object belongs to a single region, and each region belongs to a single room. 
\begin{figure*}[]
    \centering
    \includegraphics[width=1\textwidth]{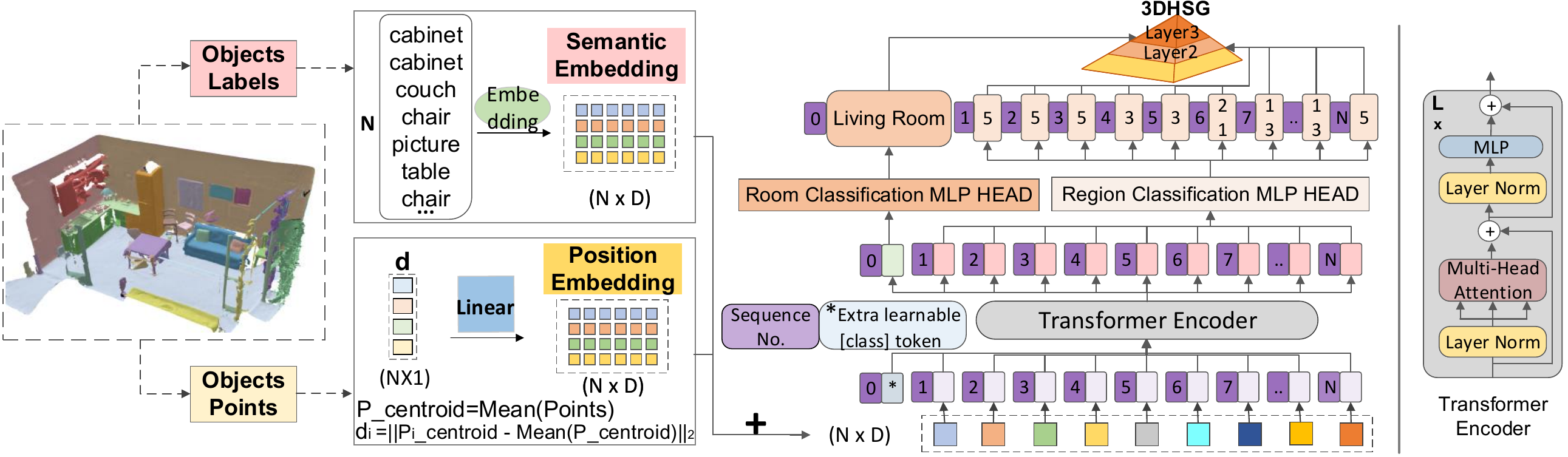}
    \caption{\textbf{TB-HSU Model Overview:} The model automatically constructs the 3DHSG for a room by completing room and region classifications, with pairs of instance-segmented point cloud and object semantic labels as inputs. The semantic embedding is derived from object labels, while position embedding is derived from object points.}
    \label{fig:model}
\end{figure*}
Additionally, each node $n \in \mathcal{N}_i$ in layer $i$ only shares edges with nodes in adjacent layers, i.e., $\mathcal{N}_{i-1}$ or $\mathcal{N}_{i+1}$. Thus, edges in the second layer connect objects to regions and regions to rooms. For any nodes $u, n \in \mathcal{N}_i$, the children of $u$ and $n$, denoted as $\mathcal{C}(n)$ and $\mathcal{C}(u)$, are disjoint, sharing no nodes or edges. 
This ensures that objects in one region are not connected to objects in another region and that regions in one room do not share edges with regions in other rooms. Details of nodes in each layer are described below:

\textbf{Layer One: Objects}.
Nodes in the object layer, $\mathcal{N}_1$, describe objects, where each node stands for an object, with an object ID, semantic label, attributes, its segments ID, and a 2D affordance vector describing both region-specific and object-specific affordances along with a list of common room categories where the object can be found.

\textbf{Layer Two: Regions}. 
Nodes in the region layer, $\mathcal{N}_2$, describe spatial regions corresponding to a region-specific affordance. Nodes contain objects with the same region-specific affordance, a region ID, contained object IDs, and a region centroid. The region centroid is calculated as the centroid of the children's object centroids. 

\textbf{Layer Three: Rooms}. 
Nodes in the room layer, $\mathcal{N}_3$, describe rooms where each node represents a room, with a room ID, scan ID, contained region IDs, and a room type.

In summary, our dataset provides a three-layered 3DHSG that describes the spatial organization of a scene with room regions defined by region-specific affordances and objects with object-specific affordances. There are 120 scene graphs with 3967 nodes and 3847 edges.

\subsection{The TB-HSU Model}
We refer to the network model that learns the 3DHSG as the TB-HSU model, and it is depicted in Figure~\ref{fig:model}. A multi-task learning framework is used to perform the combined task of room classification and specifying local spatial regions with region-specific affordances. Our model employs a transformer encoder architecture \cite{vaswani_attention_2023} that draws inspiration from the ViT~\cite{dosovitskiy_image_2021} and organizes the 3D scene data into a 1D sequence of token embeddings 
capturing both the object semantic labels and object segmented point cloud data. The 3D scene input data for the model is a ${N\times D}$ tensor, 
where $N$ represents the number of objects in the room  (zero-padded if necessary) and $D$ is the size of the token embedding.

The details for computing the token embedding for an object are now described. The object semantic labels are represented as one-hot vectors and mapped to a semantic embedding, $\textbf{E}_{\textrm{\scriptsize sem}} \in \mathbb{R}^{N \times D}$, 
using either a standard Embedding layer (or CLIP's text embedding~\cite{radford_learning_2021} in Appendices). With regard to the segmented point cloud data for each object, we calculate the centroid of the object as the mean position across all points. We then calculate a centroid for the room as the mean across all objects. We then calculate the distance between the object centroids and the centroid of the room as $d_i$, where $i$ is an index representing the i-th object. The distance data are then mapped to a position embedding, $\textbf{E}_{\textrm{\scriptsize pos}} \in \mathbb{R}^{N \times D}$, with a linear projection. The token embedding is then derived by adding together the semantic embedding and the position embedding. We also prepend a learnable embedding, $\textbf{x}_{\textrm{\scriptsize class}} \in \mathbb{R}^{1 \times D}$,\textbf{} to the token embedding in order to learn the room classification. This procedure is similar to methods used in ViT's~\cite{dosovitskiy_image_2021} and BERT's~\cite{devlin_bert_2019}. More specifically, we have as input to our transformer, $\textbf{z}_0$, which is defined as:
\begin{equation}
\textbf{z}_0 = [\textbf{x}_{\textrm{\scriptsize class}},\textbf{E}_{\textrm{\scriptsize sem}}+\textbf{E}_{\textrm{\scriptsize pos}}] \text{.}
\label{eq:1} 
\end{equation}

We follow the structure of the transformer encoder in  \cite{radford_learning_2021}, which consists of alternating layers of multiheaded self-attention (MSA) blocks and MLP blocks with a Layernorm (LN) attached before every block and a residual connection after every block. The MLP used here contains two layers with a QuickGELU non-linearity and dropout functions. The transformer network structure is shown below:

\[
\begin{array}{lll}
\textbf{z}'_l & = & \mathrm{MSA}(\mathrm{LN}(\textbf{z}_{l-1}))+\textbf{z}_{l-1} \\
\textbf{z}_l & = & \mathrm {MLP}(\mathrm{LN}(\textbf{z}'_l))+\textbf{z}'_l \, {,}
\label{eq: transformerarch}
\end{array}
\]
where layers of the transformer are indexed by the subscript~$l$. The output of the transformer, $\textbf{y}$, is computed as:
\begin{equation}
    \textbf{y} =  \mathrm{LN}(\textbf{z}_L) \, {,}
    \label{eq: transformeroutput}
\end{equation}
where $L$ is the last layer of the transformer network. Note that the output $\textbf{y}$ contains both the room and region classifications:
\begin{equation}
\textbf{y} = [\textbf{y}_{\textrm{\scriptsize room}}, \textbf{y}_{\textrm{\scriptsize region}}] \, {.}
\label{eq: outputbreakdown}
\end{equation}

Consistent with the multi-learning framework, the loss function, $\mathcal{L}(\textbf{y})$, for our network combines the loss associated with the room classification, $\mathcal{L}(\textbf{y}_{\textrm{\scriptsize room}})$, and the loss associated with the region classification,  $\mathcal{L}(\textbf{y}_{\textrm{\scriptsize region}})$, as follows:

\begin{equation}
\mathcal{L}(\textbf{y}) = \lambda \, \mathcal{L}(\textbf{y}_{\textrm{\scriptsize room}}) + (1-\lambda)\mathcal{L}(\textbf{y}_{\textrm{\scriptsize region}})\, \text{,}
\label{eq:loss}
\end{equation}
where
$\lambda = \frac{\mathcal{L}(\textbf{y}_{\textrm{\scriptsize room}})}{\mathcal{L}(\textbf{y}_{\textrm{\scriptsize room}})+\mathcal{L}(\textbf{y}_{\textrm{\scriptsize region}})}$.

\section{Experiments}
We evaluate our TB-HSU model by comparing it with two non-neural network models, three baseline neural network models, and some published models.  We run the comparisons using three different datasets.

\subsection{Datasets}
We employ the three datasets in the evaluation experiments.

\textbf{3DHSG Dataset}  We split our custom 3DHSG dataset into 96 scenes (80\%) for training and 24 scenes (20\%) for testing. We exclude object labels such as ``wall'', ``floor'', and ``ceiling'' because these regions are not annotated.

\textbf{ScanNet~\cite{dai_scannet_2017}} ScanNet is an RGB-D video dataset containing 1513 scans annotated with instance-level semantic segmentations. We split it into 1013 scenes for training and 500 scenes for testing, the same as ~\cite{huang_indoor_2020}. There are 21 different room types in the dataset. This dataset comes in two varieties: ScanNet20 with 20 object semantic labels and ScanNet200 with 200 object semantic labels.

\textbf{Matterport3D~\cite{chang_matterport3d_2017}} Matterport3D is an RGB-D dataset consisting of 90 reconstructions of indoor building-scale scenes with 2194 rooms. There are 30 room types in the dataset. We split the dataset in the same way as the benchmark and discarded rooms that contain less than 3 objects, the same as \cite{hughes_hydra_2022}.
\subsection{The Comparison Models}
\hspace{1em}\textbf{Non-Neural Network Baseline Models} We use a Random Forest (RF) classifier to perform room classification based on the object semantic labels. For the room region classification, we were inspired by two different methods:  a \emph{Term Frequency-Inverse Document Frequency (TF-IDF)} approach as described in~\cite{heikel_indoor_2022}; and a \emph{Neighbor-Vote} method~\cite{Neighbor-Vote}.

TF-IDF approach: We consider each region-specific affordance as a term; the collection of different region-specific affordances for an object as a document, and all region-specific affordances for all objects collectively as the entire set of documents. We calculate for each object a set of TF-IDF scores representing the probability that the object belongs to a particular region-specific affordance. The region-specific affordance with the highest TF-IDF score is chosen as the object's region-specific affordance.

Neighbor-Vote method: We consider the TF-IDF scores for the object and its neighboring objects when predicting the object's region-specific affordance. The close neighbors of an object are identified as having bounding boxes overlapping with the object, where the bounding boxes are determined from the object point clouds. We then calculate an object's TF-IDF score as $\alpha=0.8$ times the TF-IDF score for the object plus $(1-\alpha) = 0.2$ times the mean TF-IDF score for the close neighboring objects. As previously, the region-specific affordance with the highest score is chosen as the object's region-specific affordance.

\textbf{Neural Network Baseline Models} For the three baseline neural network models, we use the same semantic embedding and position embedding as our TB-HSU model.

MLP Model: In each layer block, we utilize two linear layers separated by a QuickGELU activation, a layer normalization, and a dropout. The number of layers matches that of the transformer layer. 

CNN Model: Six 1D convolution layers are applied along the object dimension. Batch normalization and ReLU activation are used between each convolution. 

Custom ResNet Model: We configured a ResNet \cite{ResNet} with three 1D convolutional layers, each followed by batch normalization and ReLU activation. This is followed by five residual blocks composed of two 1D convolutional layers with batch normalization and ReLU activation. 

Note that all network models are trained with the SGD optimizer, with a base learning rate of $1 \times 10^{-3}$, except for the TB-HSU model trained on ScanNet20, which uses a base learning rate of $1 \times 10^{-4}$, on a single NVIDIA GeForce GTX 3070 within 500 training epochs, except for the TB-HSU model trained on Matterport3D within 30 epochs.

\textbf{Published Reference Models} We found two published reference models for the room classification task. The second version of the Hydra~\cite{hughes_foundations_2023} model utilizes pre-trained word2vec~\cite{mikolov_efficient_2013} vectors to represent object semantic labels and concatenates them with the geometric feature vectors to perform room classification. It has been tested on the Matterport3D~\cite{chang_matterport3d_2017} dataset. A published point class histogram model (PCH)~\cite{huang_indoor_2020} has been tested on the ScanNet20 dataset. We also explore the room classification and region classification performance of GPT-4o, released by OpenAI in May 2024. According to OpenAI's official introduction, GPT-4o is the most powerful online chatbot to date. We prompt GPT-4o with figures of the scene and object semantic labels along with object centroid positions. The prompts are provided in the Appendices.

\begin{table*}[]
    \centering
    \begin{tabular}{c c | c c | c c  }
    \hline\hline
    &\textbf{Methods}& \textbf{T1 Acc\%} & \textbf{T1 mIoU\% }& \textbf{T2 Acc\% }& \textbf{T2 mIoU\% } \\
    \hline\hline
    
    \multirow{2}{*} {Non-NN} & 
   RF+TF-IDF &83.3&58.3 & 62.26&50.38\\
   &RF+Neighbor-Vote& 83.3 &58.3  & 62.62&50.48\\ \hline
    
    \multirow{3}{*} {NN}
    &MLP &29.17 & 3.24 & 44.56 $\pm$ 0.59 & 25.74 $\pm$ 0.34  \\  
    &CNN& 86.11 & 68.58  & 83.91 $\pm$ 1.14 & 72.35 $\pm$ 1.09 \\ 
    &ResNet& 87.50 & 72.02 & 85.02 $\pm$ 0.95 & 73.24 $\pm$ 0.99\\ \hline

    \multirow{2}{*}{Proposed TB-HSU} & w/o P. E. & 90.28&73.74& 84.95 $\pm$ 0.70 & 74.87 $\pm$ 1.23\\
    & with P.E. &\textbf{91.67}  & \textbf{74.60} & \textbf{87.27} $\pm$ 0.98 &\textbf{78.55} $\pm$ 2.29 \\ \hline
    &GPT-4o&\textbf{91.67}&\textbf{74.60}&44.83 &33.44\\ 
    \hline\hline
    \end{tabular}
\caption{Room and Region classification results for the 3DHSG dataset. T1 refers to the room classification task -- layer three of the 3DHSG; T2 refers to the region classification task -- layer two of the 3DHSG.}
\label{tab:Results on TB-HSU dataset}
\end{table*}
\subsection{The Performance Metrics}
We evaluate the performance of the models on the room and region classification tasks by reporting accuracy (Acc) and mean of intersection-over-union metric (mIoU), where IoU is $\frac{\mathrm{TP}}{\mathrm{TP}+\mathrm{FP}+\mathrm{FN}}$ and mIoU is average of IoU across all classes.

\section{Results}
We describe the model performance results for the room classification and region classification tasks. For the room classification task, we have the following datasets available: 3DHSG, Matterport3D, ScanNet20, and Scannet200. For the region classification task, we have only our custom 3DHSG dataset available. The TB-HSU model employs 4 transformer layers with 384 dimensions across all experiments, adapting the room classification head size (12 for 3DHSG, 30 for Matterport3D, 21 for ScanNet20 and ScanNet200), the number of kinds of object labels (191 for 3DHSG, 41 for Matterport3D, 20 for ScanNet20, and 200 for ScanNet200), and input sequence length (77 for 3DHSG, 230 for Matterport3D, 62 for ScanNet20, and 121 for ScanNet200), maintaining 7.62 $\pm$ 0.05 million parameters.

\subsection{Room Classification}
Consider now the room classification task. Table~\ref{tab:accuracy} shows results for the Matterport3D, ScanNet20, and ScanNet200 datasets. For the Matterport3D dataset, the proposed TB-HSU model obtains a performance accuracy of 62.19\% compared to the published result of 57.67\% for the Hydra model~\cite{hughes_hydra_2022}.
For the ScanNet20 dataset, the proposed TB-HSU model performs with an accuracy of 86\% or 86.8\% with or without the position embedding and better than the published result for the PCH baseline model~\cite{huang_indoor_2020}, which obtains an accuracy of 85\%. 
For the  ScanNet200 dataset, compared to ScanNet20, we find that the expanded list of object semantic labels positively impacts the TB-HSU model's performance with an increase in accuracy of 2.8\%.
\begin{table}[H]
    \centering
    \setlength{\tabcolsep}{1mm}
    \begin{tabular}{c|c c c}
    \hline \hline
    \textbf{Dataset}  & \textbf{Models} & \textbf{Inputs} & \textbf{Acc}\% \\
    \hline \hline
    \multirow{2}{*}{Matterport3D} 
    & Hydra 
        & S+P& 57.67 $\pm$ 0.57\\\cline{2-4}
    &
    Proposed TB-HSU & S+P  & \textbf{62.19} $\pm$ 0.35\\ \hline

    \multirow{4}{*}{ScanNet20} & \multirow{2}{*}{PCH} &S*& 82.8\\
                              &\scalebox{0.55}& S & 85.0\\ \cline{2-4} 

    &
    \multirow{2}{*}{Proposed TB-HSU}
    & S & 86.0\\
    && S+P & \textbf{86.8} \\    \hline
    
    \multirow{2}{*}{ScanNet200}
    &\multirow{2}{*}{Proposed TB-HSU } & S & 88.6\\
    &&S+P &\textbf{89.6}\\
    \hline\hline

   \end{tabular}
    \caption{Room Classification Accuracy Results. S refers to object semantic labels and P refers to object position data. S* refers to non-ground truth object labels. The Hydra model description can be found in~\cite{hughes_foundations_2023}. The PCH description can be found in~\cite{huang_indoor_2020}.}
     
    \label{tab:accuracy}
\end{table}

\subsection{Room and Region Classification}
In Table~\ref{tab:Results on TB-HSU dataset}, we compare model performances on the room and region classification task using our custom 3DHSG dataset. Let us consider the room classification (Task one) first. We see that the proposed TB-HSU transformer model performs better than the non-neural network models and the three baseline neural network models, obtaining a performance accuracy of 91.67\%. Note that the position encoding does not seem to strongly influence the room classification performance. This is reasonable because the room type is much more strongly dependent on the type of objects within the room rather than their position within it. Note that the GPT-4o model performs similarly to the TB-HSU model. This stands to reason because the GPT-4o model was used with manual supervision to obtain the room labels as ground truth for the 3DHSG dataset. With regards to the region classification (Task two), the proposed TB-HSU model is again the best-performing model with an accuracy of 87.27\% when using the position embedding, although the CNN and ResNet models with position embedding perform similarly when TB-HSU uses no position embedding. It is interesting to observe that the position embedding does seem to contribute some useful information for the region classification task. We note that the GPT-4o model does not perform well on the region classification task, indicating that it lacks the ability to understand the spatial concept of the object. Additional support can be found in Section Discussion.

\begin{figure*}[!h]
    \centering   
    \includegraphics[width=\textwidth]{./figures/QA.pdf}
    \caption{3DHSG from TB-HSU assist GPT-4o in a Question-Answering task to find an object not visible within the scene. \textit{Fig(a), Fig(b), Fig(c)} are inserted appropriately place within the prompts.}
    \label{fig:QA}
\end{figure*}

\section{Discussion}
\label{sec:discussion}
The capability of large language models (LLMs) is increasing at a rapid rate. For example, previous methods such as ConceptFusion~\cite{jatavallabhula_conceptfusion_2023}, SayCan~\cite{ahn_as_2022}, and 3D-LLM~\cite{hong_3d-llm_2023} have demonstrated the viability of large language models as knowledge bases that can be queried for generating task-level plans. Along these lines, we suggest that the TB-HSU model and its resulting 3DHSG may be useful as a prompt input to LLMs. Consider the interesting case of finding an object that is not visible in the scene. Figure~\ref{fig:QA} illustrates a scenario from the 3RScan~\cite{wald_rio_2019} interactive household simulator. The system is tasked with finding an object of interest (shampoo) that is not contained within the map because it is concealed within a receptacle or too small to detect. We investigate the behavior of  GPT-4o. As seen in Figure~\ref{fig:QA}, GPT-4o, when provided only with an image from an instance-segmented point cloud, struggles to identify objects in the scene. However, with labeled point clouds and object semantic labels, objects can be clearly identified but there is still difficulty in suggesting where to find the missing object. When we include the 3DHSG that results from the TB-HSU model, we find that the outputs from the GPT-4o model are much more reasonable.

\section{Conclusion}
In conclusion, this study on Hierarchical 3D Scene Understanding with Contextual Affordances demonstrates that the affordances of objects vary with different levels of spatial context. We introduce the 3DHSG dataset annotated with region-specific and object-specific affordances and organized into three spatial layers: Objects, Regions, and Rooms. We propose the TB-HSU model for solving the multi-task problem of room classification and region classification, with a promising performance over multiple baselines. The TB-HSU model produces a 3D hierarchical scene graph that is useful for evaluating task objectives based on a spatial and functional framework that allows affordances to vary with the spatial context. Additionally, the spatial organization of the 3DHSG dataset enhances the performance of large language models (LLMs) in question-answering tasks. In future work, we will improve our dataset and model, e.g., reducing the variety of objects and distinguishing similar regions.

\section{Acknowledgments}
We would like to thank the Australian government for their funding support via a CRC Projects Round 11 grant.

\bibliography{TB-HSU}
\clearpage
\newpage
\section{Appendices}

\subsection{3DHSG Dataset}
\begin{figure*}[h]
    \centering
        \includegraphics[width=\textwidth]{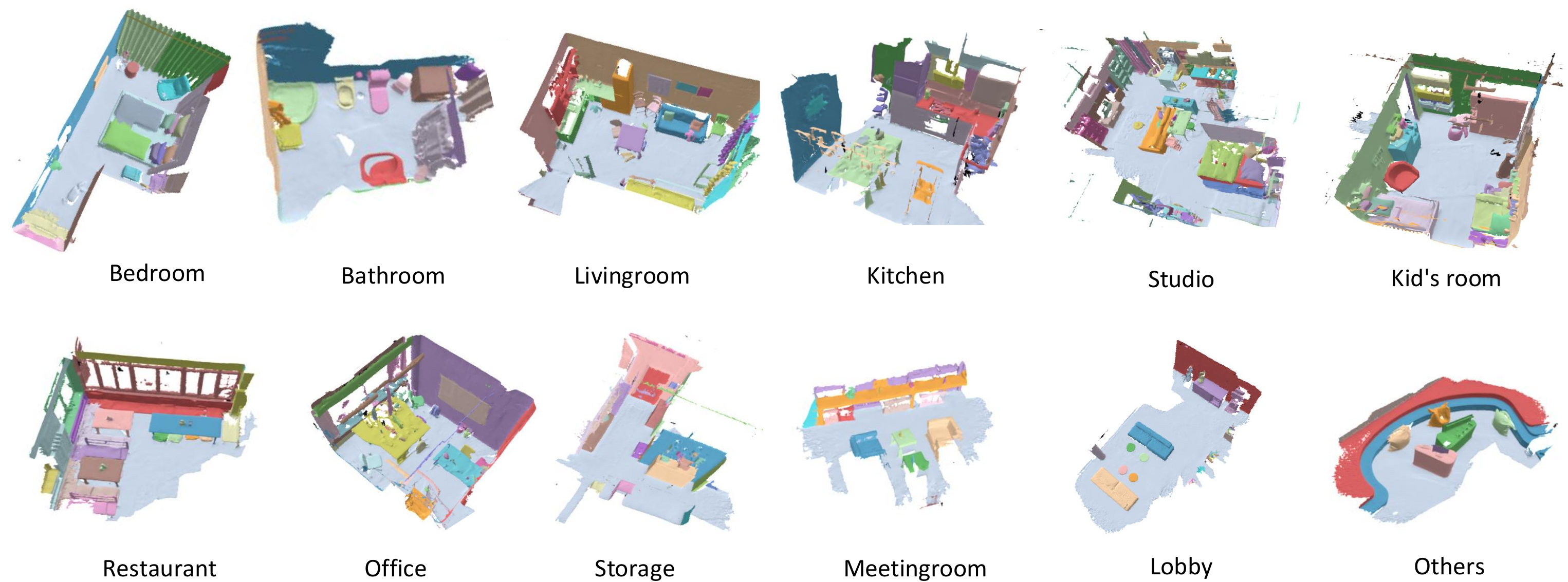}
    \caption{Different room categories in 3DHSG dataset}
    \label{fig:DifferentSceneType}
\end{figure*}

\begin{figure}[h]
    \centering
    \includegraphics[width=\columnwidth]{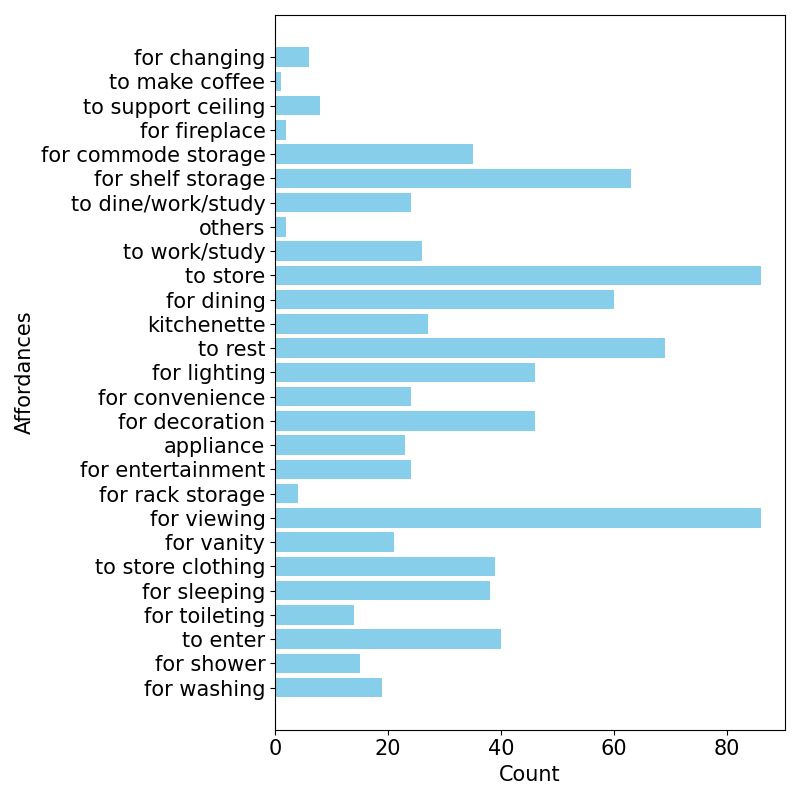}
    \caption{Reigion-specific affordances distribution in 3DHSG dataset}
    \label{fig:AFFlayer2}
\end{figure}

The 3DHSG dataset extends the 3DSSG dataset, which extends the 3RScan dataset. It contains 120 3DHSGs that represent the spatial organization of a scene using a three-layered graph with nodes containing: Objects, Regions, and Rooms. The 3DHSG dataset has twelve different room types as shown in Figure~\ref{fig:DifferentSceneType}; twenty-seven different types of room regions defined by unique region-specific affordances as shown in Figure~\ref{fig:AFFlayer2}; 
and 196 kinds of objects, as shown in Table ~\ref{tab:objlab}, while object-specific affordances are available in our public dataset. Some examples in 3DHSG are shown in Figure~\ref{fig:datasetEx_1}, Figure~\ref{fig:datasetEx_2}, and Figure~\ref{fig:datasetEx_3}.


\begin{figure*}[h]
    \centering
    \includegraphics[width=\textwidth]{./figures/datasetEx_1.pdf}
    \caption{Example in 3DHSG dataset (more details in digital files)}
    \label{fig:datasetEx_1}
\end{figure*}
\begin{figure*}[h]
    \centering
    \includegraphics[width=\textwidth]{./figures/datasetEx_2.pdf}
    \caption{Example in 3DHSG dataset (more details in digital files)}
    \label{fig:datasetEx_2}
\end{figure*}
\begin{figure*}[h]
    \centering
    \includegraphics[width=\textwidth]{./figures/datasetEx_3.pdf}
    \caption{Example in 3DHSG dataset (more details in digital files)}
    \label{fig:datasetEx_3}
\end{figure*}

\begin{table*}[ht]
\centering
\begin{tabular}{|l|l|l|l|l|}
\hline
Air Conditioner & Armchair & Baby Bed & Baby Changing Unit & Baby Seat \\ \hline
Backpack        & Bag      & Balcony  & Bar               & Basket    \\ \hline
Bath Cabinet    & Bathtub  & Bean Bag & Bed               & Bench     \\ \hline
Bidet           & Bin      & Blanket  & Blinds            & Board     \\ \hline
Boiler          & Book     & Books    & Bookshelf         & Boots     \\ \hline
Bottle          & Box      & Boxes    & Bucket            & Cabinet   \\ \hline
Candle          & Carpet   & Cart     & Ceiling           & Chair     \\ \hline
Cleaning Brush  & Cleanser & Clock    & Closet            & Clothes   \\ \hline
Clothes Dryer   & Clutter  & Coffee Machine & Coffee Table & Column    \\ \hline
Commode         & Container & Couch    & Couch Table       & Counter   \\ \hline
Cube            & Cup      & Cupboard & Curtain           & Cushion   \\ \hline
Cutting Board   & Decoration & Desk    & Device            & Dishes    \\ \hline
Dispenser       & Door     & Door /Other Room & Doorframe  & Drawer    \\ \hline
Dressing Table  & Exhaust Hood & Extractor Fan & Fan       & Fireplace \\ \hline
Floor           & Flower   & Folder   & Frame             & Furniture \\ \hline
Hair Dryer      & Handbag  & Hanger   & Hanging Cabinet   & Heater    \\ \hline
Humidifier      & Item     & Items    & Juicer            & Kettle    \\ \hline
Kids Bicycle    & Kitchen Appliance & Kitchen Cabinet & Kitchen Counter & Kitchen Hood \\ \hline
Kitchen Item    & Kitchen Towel & Ladder & Lamp            & Laptop    \\ \hline
Laundry Basket  & Letter   & Light    & Loft Bed          & Luggage   \\ \hline
Magazine Files  & Magazine Rack & Menu & Microwave        & Mirror    \\ \hline
Monitor         & Napkins  & Nightstand & Object          & Organizer \\ \hline
Ottoman         & Oven     & Pack     & Painting          & Pan       \\ \hline
Paper           & Paper Holder & Paper Towel Dispenser & Papers & PC \\ \hline
Photo Frame     & Photos   & Picture  & Pile Of Books     & Pile Of Candles \\ \hline
Pillar          & Pillow   & Plant    & Planter           & Plate     \\ \hline
Player          & Podest   & Poster   & Pot               & Price Tag \\ \hline
Printer         & Puf      & Puppet   & Rack              & Radio     \\ \hline
Rail            & Refrigerator & Rolling Pin & Round Table & Rug      \\ \hline
Salt            & Shelf    & Shoe Commode & Shoe Rack     & Shoes     \\ \hline
Showcase        & Shower   & Shower Curtain & Shower Wall & Side Table \\ \hline
Sign            & Sink     & Soap Dispenser & Sofa        & Stairs    \\ \hline
Stand           & Stool    & Storage  & Storage Container & Stove     \\ \hline
Stuffed Animal  & Suitcase & Table    & Table Lamp        & Telephone \\ \hline
Toilet          & Toilet Paper & Toilet Paper Dispenser & Toilet Paper Holder & Toiletry \\ \hline
Towel           & Trash Can & Trashcan & Treadmill       & TV        \\ \hline
TV Stand        & Vacuum Cleaner & Vase & Wall          & Wall /Other Room \\ \hline
Wall Frame      & Wardrobe & Wardrobe Door & Washbasin    & Washing Machine \\ \hline
Water Heater    & Whiteboard & Window & Windowsill       & Wood Box  \\ \hline
Xbox            &          &          &                  &           \\ \hline
\end{tabular}
\caption{List of Object Labels}
\label{tab:objlab}
\end{table*}

\subsection{Ablation Study}

\subsubsection{Replacement of Semantic Embedding with Text Embedding from pre-trained CLIP} In the work above, the semantic embedding was trained using only the object semantic labels available in the provided dataset. In this study, we replace the semantic embedding layer with text embedding from a pre-trained CLIP, which has 49,408 words in its vocabulary. More specifically, we employ the text encoder from the pre-trained `ViT-B/32' CLIP variant. The size of the text embedding is reduced from 512 to 384 using matrix multiplication via a learnable matrix. 

Results are shown in Table~\ref{tab:clipvoc}. The model exhibits a decline in performance for region classification compared to the previous method; however, it demonstrates an improved mIoU in the room classification task.

\begin{table}[H]
\centering
\small
\setlength{\tabcolsep}{0.8mm}
\begin{tabular}{c c c c c c}
\hline
     \textbf{Method} & \textbf{T1 Acc\%} & \textbf{T1 mIoU\%}& \textbf{
     T2 Acc\%}& \textbf{
     T2 mIoU\%}\\ \hline
     TB-HSU-T.E.&91.67&79.63&83.86&76.45
     \\

\hline
\end{tabular}

\caption{The TB-HSU model utilizes the text encoder from the pre-trained CLIP model (`ViT-B/32') as the text embedding, replacing the semantic embedding.}

\label{tab:clipvoc}

\end{table}

\subsection{Prompts to GPT-4o}
\label{sec:prompts}
\begin{figure}[h]
    \centering
    \begin{subfigure}
        \centering
        \includegraphics[width=\columnwidth]{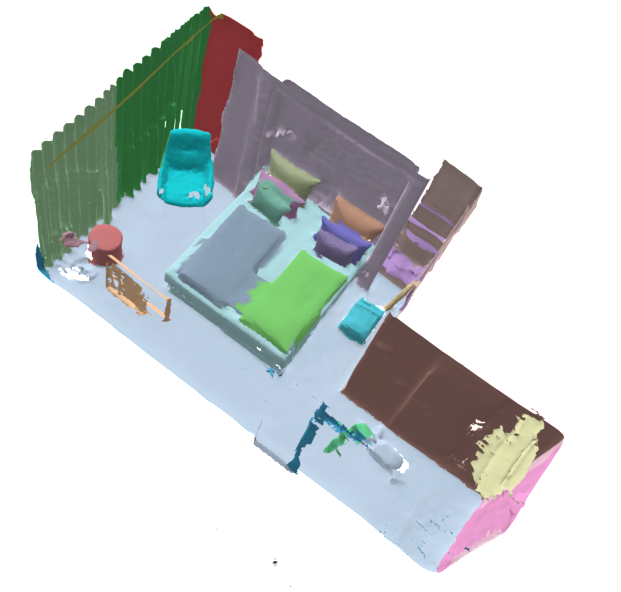}
        \caption{Details of fig(1) in prompts}
        \label{fig:figa}
    \end{subfigure}
    \begin{subfigure}
        \centering
        \includegraphics[width=\columnwidth]{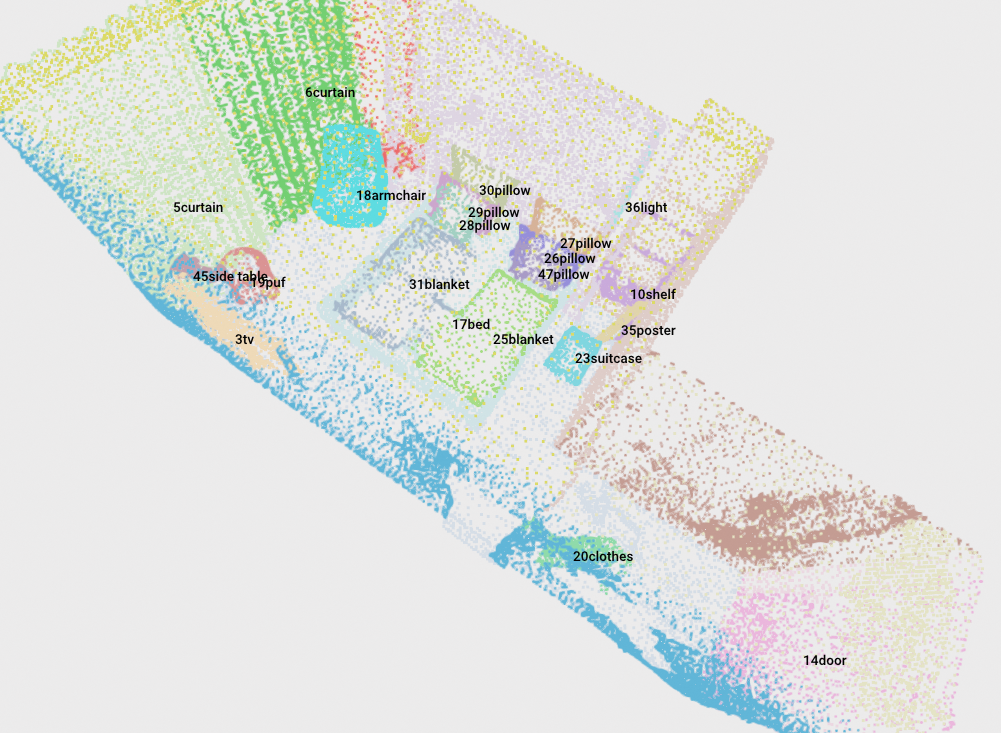}
        \caption{Details of fig(2) in prompts}
        \label{fig:figb}
    \end{subfigure}
\end{figure}

GPT-4o is OpenAI's new flagship model that can reason across audio, vision, and text in real-time, according to their official website. To ensure a fair comparison of our method and GPT-4o, we validate each scene from the dataset individually using figures of instance-segmented point clouds and object-annotated point clouds, with a comprehensive list of room categories and region-specific affordances. We provide examples of the GPT-4o prompts used on the OpenAI Playground (\url{https://platform.openai.com/playground/chat?models=gpt-4o}), with figures inserted in the corresponding place of the prompts, and details shown in Figure~\ref{fig:figa}, \ref{fig:figb}. Furthermore, we use the same prompts to GPT-4o in Section 3.1.

\begin{lstlisting}
I am in a room with an instance-segmented point cloud, 
displayed in fig(1). The Object IDs and Labels are shown 
in fig(2). Here are the objects with their labels and 
respective positions:
3tv [-1.8646073   0.24237497  0.0808332 ]
5curtain [-2.5952835   1.5693542  -0.15880549]
6curtain [-1.6328962   2.8131962   0.00821053]
10shelf [ 1.4300133  1.1042987 -0.5555024]
14door [ 1.9846358  -2.025839   -0.12992856]
17bed [-0.27750984  0.94435805 -0.941797  ]
18armchair [-1.2530015  2.372524  -0.8003811]
19puf [-2.2171662   1.3928659  -0.98535466]
20clothes [ 0.36916748 -1.3091931   0.20796768]
23suitcase [ 0.99524784  0.68045133 -1.1444317 ]
25blanket [ 0.08170674  0.6514464  -0.6796305 ]
26pillow [ 0.5986713  1.472307  -0.5391769]
27pillow [ 0.7603667   1.6041864  -0.46824417]
28pillow [-0.21387196  1.8667188  -0.5851146 ]
29pillow [-0.12313294  2.0143044  -0.5656808 ]
30pillow [-0.01137643  2.2150292  -0.46969303]
31blanket [-0.6965012  1.2289908 -0.685415 ]
35poster [ 1.4284203   0.8868982  -0.89788866]
36light [1.0646423 1.3352208 0.3758623]
45side table [-2.572783   1.2327604 -0.6531584]
47pillow [ 0.5339395   1.3031684  -0.56844467]

Question:
Based on the objects in this room, what is the common 
type for this room from the following list: 
["bedroom", "bathroom", "livingroom", "kitchen", 
"studio", "kidsroom", "restaurant", "office", "storage",
"meetingroom", "lobby", "others"]? Considering the 
region-specific affordance, what is the 
individual affordance for each object from the 
following list: ["for washing", "for shower", "to enter", 
"for toileting",  "for sleeping", "to store clothing", 
"for vanity", "for viewing", "to store shoes", 
"for entertainment", "appliance", "for decoration", 
"for convenience", "for lighting", "to rest", 
"kitchenette", "for dining", "to store", "to work/study", 
"others", "to dine/work/study", "for shelf storage", 
"for commode storage", "for fireplace", "to support ceiling",
 "to make coffee", "for changing"]?

Please provide the answers in JSON format like this:
{
  "Layer1": "common_type",
  "Layer2": {
    "Object ID": "individual_affordance",
    "Object ID": "individual_affordance",
    ...
  }
}
\end{lstlisting}
We then reconstruct the answers to maintain consistent description and follow the same accuracy and mIoU calculation methods to obtain the performance results, as shown in Table~\ref{tab:Results on TB-HSU dataset}.
\end{document}